\documentclass[11pt,a4paper]{article}
\usepackage{geometry}
\usepackage{graphicx}
\usepackage{amsmath,amssymb,amsfonts}
\usepackage{amsthm}
\usepackage{booktabs}
\usepackage{multirow}
\usepackage{array}
\usepackage{colortbl}
\usepackage{caption}
\usepackage{subcaption}
\usepackage{xcolor}
\usepackage{listings}
\usepackage{algorithm}
\usepackage{algpseudocode}
\usepackage{hyperref}
\usepackage{enumitem}
\usepackage{fancyhdr}
\usepackage{setspace}
\usepackage{tikz}
\usepackage{pgfplots}
\usepackage{placeins}
\usepackage{capt-of}
\pgfplotsset{compat=1.18}
\usetikzlibrary{shapes.geometric, arrows.meta, positioning, calc, fit, backgrounds}
\definecolor{primaryblue}{RGB}{31,119,180}
\definecolor{accentgreen}{RGB}{44,160,44}
\definecolor{warnred}{RGB}{214,39,40}
\definecolor{neutralgray}{RGB}{127,127,127}
\definecolor{lightbg}{RGB}{248,249,250}
\definecolor{bordergray}{RGB}{220,220,220}
\definecolor{focusbg}{RGB}{237,242,255}
\definecolor{bestbg}{RGB}{232,245,233}
\definecolor{tradebg}{RGB}{255,248,230}
\hypersetup{
    colorlinks=true,
    linkcolor=primaryblue,
    citecolor=primaryblue,
    urlcolor=primaryblue
}
\newcommand{\teamlogo}{%
    \IfFileExists{figures/logo.png}{%
        \raisebox{-0.2em}{\includegraphics[height=1.2em]{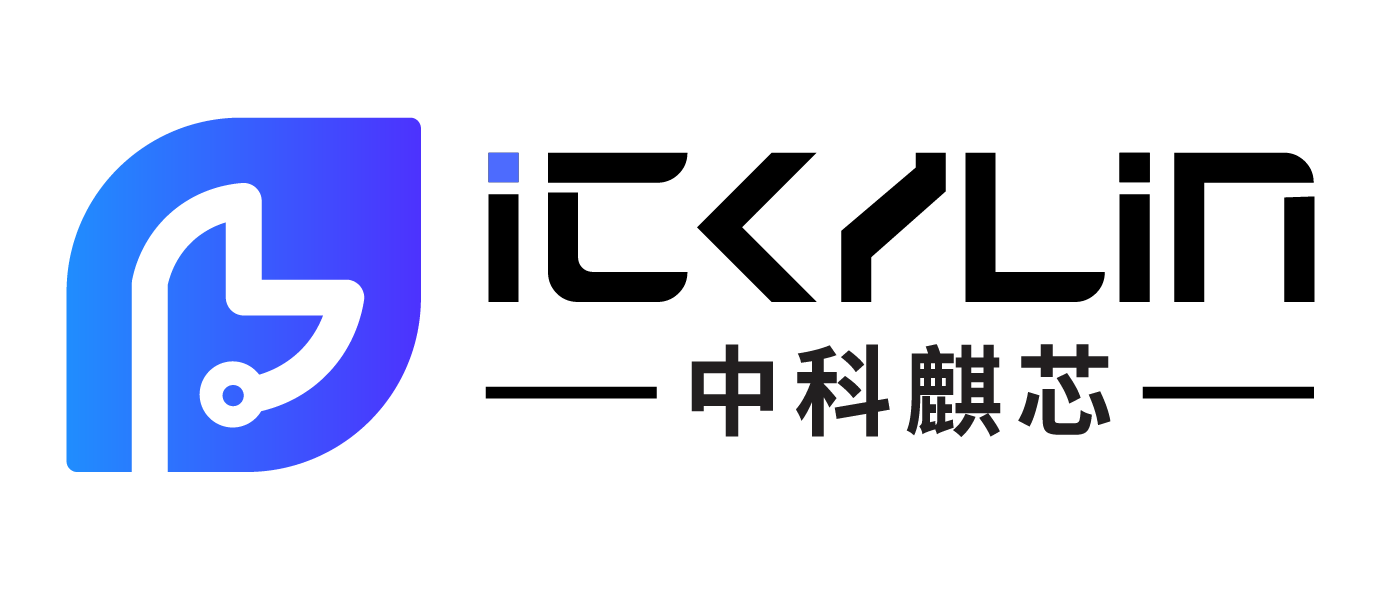}}\quad%
    }{}%
}
\theoremstyle{definition}
\newtheorem{definition}{\textbf{Definition}}[section]
\setlist[enumerate]{leftmargin=2em,itemsep=2pt}
\setlist[itemize]{leftmargin=2em,itemsep=2pt}
\newcommand{\bestcell}[1]{\cellcolor{bestbg}\textbf{#1}}
\title{\textbf{ChipLingo: A Systematic Training Framework for Large Language Models in EDA}}
\author{
Lei Li \quad Xingwen Yu \quad Jianguo Ni \quad Junxuan Zhu\\
Jieqiong Zhang \quad Jian Zhao \quad Zhi Liu\\
\textbf{Ickylin AI Team}\\[0.3em]
\texttt{info@ickylin.com}\\
\url{https://www.ickylin.com}
}
\date{}
\begin{document}
\maketitle
\thispagestyle{fancy}
\begin{abstract}
\noindent
With the rapid advancement of semiconductor technology, the complexity of Electronic Design Automation (EDA) tools has increased substantially, making EDA a highly knowledge-intensive and document-driven engineering domain. Although large language models (LLMs) have demonstrated strong capabilities in general tasks, their direct application to the EDA domain still faces significant challenges, including insufficient domain expertise, cross-tool knowledge confusion, and degradation of retrieval-augmented generation (RAG) capabilities after domain training. To address these challenges, this paper presents ChipLingo, a systematic training pipeline for domain-adapted large language models tailored to EDA scenarios.

The proposed pipeline consists of three stages. First, we construct domain-specific corpora through multi-source data curation and question-answering (QA) augmentation. Second, during the domain-adaptive pretraining phase, we compare different parameter training strategies and examine their effects on domain adaptation performance and selected general capability benchmarks. Third, through instruction alignment and RAG scenario training targeting diverse retrieval conditions, we enhance the model's ability to leverage external knowledge. To evaluate model performance on EDA question-answering tasks, we curate an internal benchmark called EDA-Bench, covering multiple representative EDA tool scenarios, which is planned for public release.

Experimental results demonstrate that ChipLingo-8B achieves 59.7\% accuracy on EDA-Bench, outperforming the base model of the same scale and surpassing some larger general-purpose models. ChipLingo-32B reaches 70.02\% accuracy, approaching the performance of leading closed-source commercial models on this benchmark. Furthermore, we observe that under the current experimental settings, QA augmentation contributes to improved domain performance, Partial FT achieves a better trade-off between domain adaptation and general capability retention compared to LoRA, and explicit RAG scenario training can mitigate the decline in retrieval information utilization that occurs after domain training. These results indicate that a systematic domain training pipeline provides practical value for enhancing LLM performance on knowledge-intensive EDA tasks, while also laying the foundation for building EDA agents or harness capabilities that rely on external knowledge.

\vspace{0.5em}
\end{abstract}
%
\section{Introduction}
\label{sec:intro}

\subsection{Background}

With the rapid advancement of semiconductor technology, the scale and complexity of modern chip systems continue to grow. State-of-the-art chips now contain billions or even tens of billions of transistors, with design workflows spanning multiple stages including logic synthesis, physical design, timing analysis, verification, and power optimization. To address these challenges, Electronic Design Automation (EDA) tools have become indispensable infrastructure~\cite{he2023chateda}.

However, the EDA ecosystem exhibits substantial complexity. Different design stages typically rely on distinct specialized tools, each providing extensive command interfaces, parameter configurations, and design constraint mechanisms. In practice, engineers must frequently consult large volumes of technical documentation, user manuals, and community forums to understand tool behavior, debug design issues, and optimize design flows. Moreover, significant differences exist among EDA tools in terms of command syntax, design workflows, and configuration approaches, further raising the learning curve and barriers to entry~\cite{xu2024chipexpert}.

\textbf{Knowledge-Intensive Characteristics.} In engineering practice, substantial knowledge exists in the form of documentation, accumulated experience, and question-answering exchanges, distributed across diverse sources. This knowledge-intensive characteristic makes EDA a typical ``document-driven'' engineering domain. For more sophisticated EDA intelligent assistants, agents, or harness systems, models must not only answer questions but also maintain stable knowledge alignment across external documents, tool feedback, and task workflows. Therefore, reliable retrieval utilization capability can be viewed as a foundational ability for deploying higher-level EDA intelligent systems.

\subsection{Problem Definition and Challenges}

\subsubsection{Limitations of General LLMs in the EDA Domain}

Large Language Models (LLMs) have achieved significant progress in natural language understanding, code generation, and knowledge-based question answering. General-purpose LLMs have demonstrated strong knowledge representation and task generalization capabilities across multiple domains~\cite{lewis2020rag}. Nevertheless, directly applying general LLMs to answer EDA domain questions still faces notable limitations:
\begin{itemize}
    \item \textbf{Insufficient Domain Expertise}: General models typically lack adequate EDA-specific knowledge, leading to inaccurate or overly generic responses to domain-specific questions.
    \item \textbf{Cross-Tool Knowledge Confusion}: Due to significant differences in command interfaces and design constraints across EDA tools, models may confuse terminology and usage patterns between different tools.
    \item \textbf{Lack of Real-Time Knowledge}: EDA domain questions often require reference to specific documentation or tool version information, necessitating the ability to retrieve up-to-date knowledge.
\end{itemize}

\subsubsection{RAG Capability Degradation After Domain Training}

Given that EDA domain knowledge updates frequently and relies heavily on documentation, Retrieval-Augmented Generation (RAG) is considered an important technical approach for addressing domain knowledge challenges~\cite{lewis2020rag,gao2023rag,karakurt2026rag}. However, after domain-specific continued training, model RAG capabilities may exhibit notable degradation. After acquiring domain knowledge, models tend to rely more on parametric knowledge for answering questions rather than utilizing retrieved external knowledge. This phenomenon is particularly pronounced in knowledge-intensive domains and poses new challenges for RAG system stability. Furthermore, this issue affects not only single-turn question-answering scenarios but also directly impacts EDA agent or harness systems that depend on external documents, tool feedback, and contextual state for decision-making.

\subsection{Main Contributions}

This paper presents \textbf{ChipLingo}, developed around knowledge-intensive EDA question-answering scenarios. The main contributions include:
\begin{enumerate}
    \item We construct a systematic LLM training pipeline for the EDA domain, covering data preparation, domain-adaptive pretraining, instruction alignment, and RAG scenario training.
    \item Under the current experimental settings, we systematically compare the effects of QA augmentation, LoRA, Full FT, and Partial FT training strategies on EDA domain adaptation and selected general capability benchmarks.
    \item We observe and quantify changes in model retrieval information utilization after domain training, demonstrating that RAG scenario training targeting diverse retrieval conditions can mitigate this issue on the current benchmark.
    \item We curate EDA-Bench, a benchmark covering multiple representative EDA tool scenarios for evaluating model performance on EDA engineering question-answering tasks. This benchmark is currently undergoing further curation and preparation for release.
\end{enumerate}

\section{Related Work}
\label{sec:related}

\subsection{Domain-Specific Large Language Models}

With the successful application of LLMs in natural language processing tasks, an increasing number of studies have explored how to build domain-specific large language models~\cite{colombo2024saullm,xu2026iscript}. These works demonstrate that continued pretraining or task adaptation on domain-specific corpora can significantly improve model performance in specialized scenarios.

In the EDA and chip design domain, prior works have explored domain-specific models for chip design scenarios. ChipNeMo~\cite{liu2023chipnemo} systematically investigated domain-adaptive tokenizers, continued pretraining, instruction alignment, and retrieval augmentation for chip design. Additionally, ChipExpert~\cite{xu2024chipexpert} and customized RAG/benchmark work for EDA documentation QA~\cite{pu2024eda} further demonstrate that general LLMs require domain adaptation to adequately handle EDA tasks. Meanwhile, VerilogEval~\cite{thakur2023verilogeval} and similar works have constructed more focused evaluation frameworks from the perspective of hardware code generation, indicating that benchmarks and application scenarios for chip design tasks are becoming increasingly refined.

Compared to these works, the EDA domain exhibits more complex knowledge structures. EDA knowledge exists not only in technical documentation but is also distributed across engineering experience, tool commands, and design workflows in various knowledge forms.

\subsection{Domain Adaptation and Continued Pretraining}

Domain Adaptation is an important approach for building domain-specific models~\cite{tianjun2024raft,soudani2024rag,colombo2024saullm}. A common method involves performing continued pretraining on domain corpora based on general foundation models to enhance model understanding of domain knowledge~\cite{xu2026iscript,colombo2024saullm,tianjun2024raft}.

Inspired by research such as RAFT~\cite{tianjun2024raft}, this paper explores a QA-augmented domain continued pretraining strategy. Unlike traditional approaches that use only documents for pretraining, we introduce question-answering format data during the pretraining phase, enabling the model to learn the relationship between knowledge and tasks earlier in training.

\subsection{Parameter-Efficient Fine-Tuning Methods}

As large language model sizes continue to grow, the computational cost of full-parameter fine-tuning has become a significant limiting factor for model deployment~\cite{zhang2024adalore,dettmers2023qlora}. Researchers have proposed a series of parameter-efficient fine-tuning methods, including Adapter~\cite{houlsby2019adapter}, Prefix-Tuning~\cite{li2021prefix}, and LoRA~\cite{hu2021lora}.

However, recent research indicates that parameter-efficient fine-tuning methods may have certain limitations in knowledge-intensive tasks~\cite{pletenev2025lora}. Due to constraints on the expressiveness of low-rank updates, models may struggle to fully absorb domain knowledge when tasks require learning large amounts of fine-grained knowledge.

\subsection{Retrieval-Augmented Generation}

Retrieval-Augmented Generation (RAG) has emerged as an important method for enhancing LLM knowledge utilization capabilities in recent years~\cite{lewis2020rag,asai2023selfrag,edge2024graphrag,sarmah2024hybridrag}. Researchers have proposed various approaches to improve RAG, including Self-RAG~\cite{asai2023selfrag}, GraphRAG~\cite{edge2024graphrag}, and HybridRAG~\cite{sarmah2024hybridrag}.

In this study, we further observe an important phenomenon: after domain training, model RAG capabilities may exhibit notable degradation. This phenomenon is particularly pronounced in knowledge-intensive domains and poses new challenges for RAG system design.

\subsection{Distinctions from Existing Work}

Compared to existing work, the main distinctions of this paper include:
\begin{itemize}
    \item We propose a systematic LLM training framework for the EDA domain;
    \item We explore the role of QA-augmented pretraining in domain model training;
    \item We systematically analyze the limitations of parameter-efficient fine-tuning in knowledge-intensive domains;
    \item We investigate the phenomenon of RAG capability degradation during domain training and propose corresponding training strategies.
\end{itemize}

Furthermore, this paper views retrieval information utilization capability as a critical foundational ability for higher-level EDA intelligent systems. In other words, the RAG grounding problem addressed in this paper relates not only to single-turn question-answering quality but also directly determines whether subsequent agent or harness systems can stably leverage external knowledge and tool feedback in real engineering workflows.

\section{Method}
\label{sec:method}

\subsection{Problem Formalization}

Before introducing the ChipLingo training framework, we first provide formal definitions for EDA domain question-answering tasks and the model training process.

\begin{definition}[EDA Question-Answering Task]
Given an EDA domain question $q$ and an optional set of retrieved documents $C = \{c_1, c_2, \ldots, c_k\}$, a large language model $M_\theta$ is required to generate an answer sequence $a = M_\theta(q, C)$ such that $a$ is semantically equivalent to the ground-truth answer $a^*$. When $C = \emptyset$, the model relies solely on parametric knowledge for answering; when $C \neq \emptyset$, the model must incorporate retrieved context to generate the answer.
\end{definition}

\begin{definition}[RAG Gain Rate]
Let $C_i^+$ denote the retrieved context containing the correct answer. The RAG gain rate is defined as the improvement in model accuracy when provided with correct retrieved context:
\begin{equation}
\Delta_{\text{rag}} = \frac{1}{N} \sum_{i=1}^{N} \mathbb{I}\bigl(M_\theta(q_i, C_i^+) = a_i^*\bigr) - \frac{1}{N} \sum_{i=1}^{N} \mathbb{I}\bigl(M_\theta(q_i, \emptyset) = a_i^*\bigr)
\end{equation}
where $a_i^*$ is the ground-truth answer and $\mathbb{I}(\cdot)$ is the indicator function. A higher $\Delta_{\text{rag}}$ indicates that the model can more effectively utilize correct retrieved information.
\end{definition}

\begin{definition}[Noise Impact Degree]
Let $C_i^-$ denote retrieved context irrelevant to question $q_i$. The noise impact degree is defined as the change in model accuracy when irrelevant retrieved context is introduced, relative to the no-retrieval condition:
\begin{equation}
\Delta_{\text{noise}} = \frac{1}{N} \sum_{i=1}^{N} \mathbb{I}\bigl(M_\theta(q_i, C_i^-) = a_i^*\bigr) - \frac{1}{N} \sum_{i=1}^{N} \mathbb{I}\bigl(M_\theta(q_i, \emptyset) = a_i^*\bigr)
\end{equation}
This metric measures the degree of interference caused by irrelevant retrieved context on model performance. A $\Delta_{\text{noise}}$ closer to 0 indicates stronger model resistance to external noise.
\end{definition}

Based on the above definitions, the three-stage training objectives of ChipLingo can be formalized as:
\begin{align}
\theta_1 &= \arg\min_{\theta} \mathcal{L}_{\text{PT}}\bigl(M_\theta, D_{\text{pt}}\bigr) \label{eq:dap}\\
\theta_2 &= \arg\min_{\theta} \mathcal{L}_{\text{SFT}}\bigl(M_{\theta_1}, D_{\text{sft}}\bigr) \label{eq:sft}\\
\theta^* &= \arg\min_{\theta} \mathcal{L}_{\text{RAG}}\bigl(M_{\theta_2}, D_{\text{rag}}\bigr) \label{eq:rag}
\end{align}

Specifically, the loss functions for each stage are defined as follows:
\begin{itemize}
    \item \textbf{Domain-Adaptive Pretraining Stage}: The standard causal language modeling loss is employed, maximizing the log-likelihood over the pretraining corpus $D_{\text{pt}}$:
    \begin{equation}
    \mathcal{L}_{\text{PT}} = -\mathbb{E}_{(x) \sim D_{\text{pt}}} \left[ \sum_{t=1}^{|x|} \log P_{M_\theta}\bigl(x_t \mid x_{<t}\bigr) \right]
    \end{equation}
    \item \textbf{Instruction Alignment Stage}: Supervised fine-tuning loss is applied, optimizing the generation probability of final answers based on question-answer pairs $(q, a) \in D_{\text{sft}}$. Auxiliary reasoning annotations, when available in the raw instruction data, are not used as explicit generation targets:
    \begin{equation}
    \mathcal{L}_{\text{SFT}} = -\mathbb{E}_{(q, a) \sim D_{\text{sft}}} \left[ \sum_{t=1}^{|a|} \log P_{M_\theta}\bigl(a_t \mid q, a_{<t}\bigr) \right]
    \end{equation}
    \item \textbf{RAG Scenario Fine-Tuning Stage}: Under conditions that include retrieved context $C$, the answer generation probability is optimized using training data $(q, C, a) \in D_{\text{rag}}$:
    \begin{equation}
    \mathcal{L}_{\text{RAG}} = -\mathbb{E}_{(q, C, a) \sim D_{\text{rag}}} \left[ \sum_{t=1}^{|a|} \log P_{M_\theta}\bigl(a_t \mid q, C, a_{<t}\bigr) \right]
    \end{equation}
\end{itemize}

For the \textbf{partial parameter training strategy}, let the complete model parameters be $\theta$. We freeze the parameters of several bottom layers, denoted as $\theta_{\text{freeze}} \subset \theta$, and update only a selected subset of the remaining parameters $\theta_{\text{train}} \subset \theta \setminus \theta_{\text{freeze}}$. During training, only $\theta_{\text{train}}$ is updated:
\begin{equation}
\theta_{\text{train}}^{(t+1)} = \theta_{\text{train}}^{(t)} - \eta \nabla_{\theta_{\text{train}}} \mathcal{L}
\end{equation}

\begin{figure}[htbp]
    \centering
    \includegraphics[width=\textwidth]{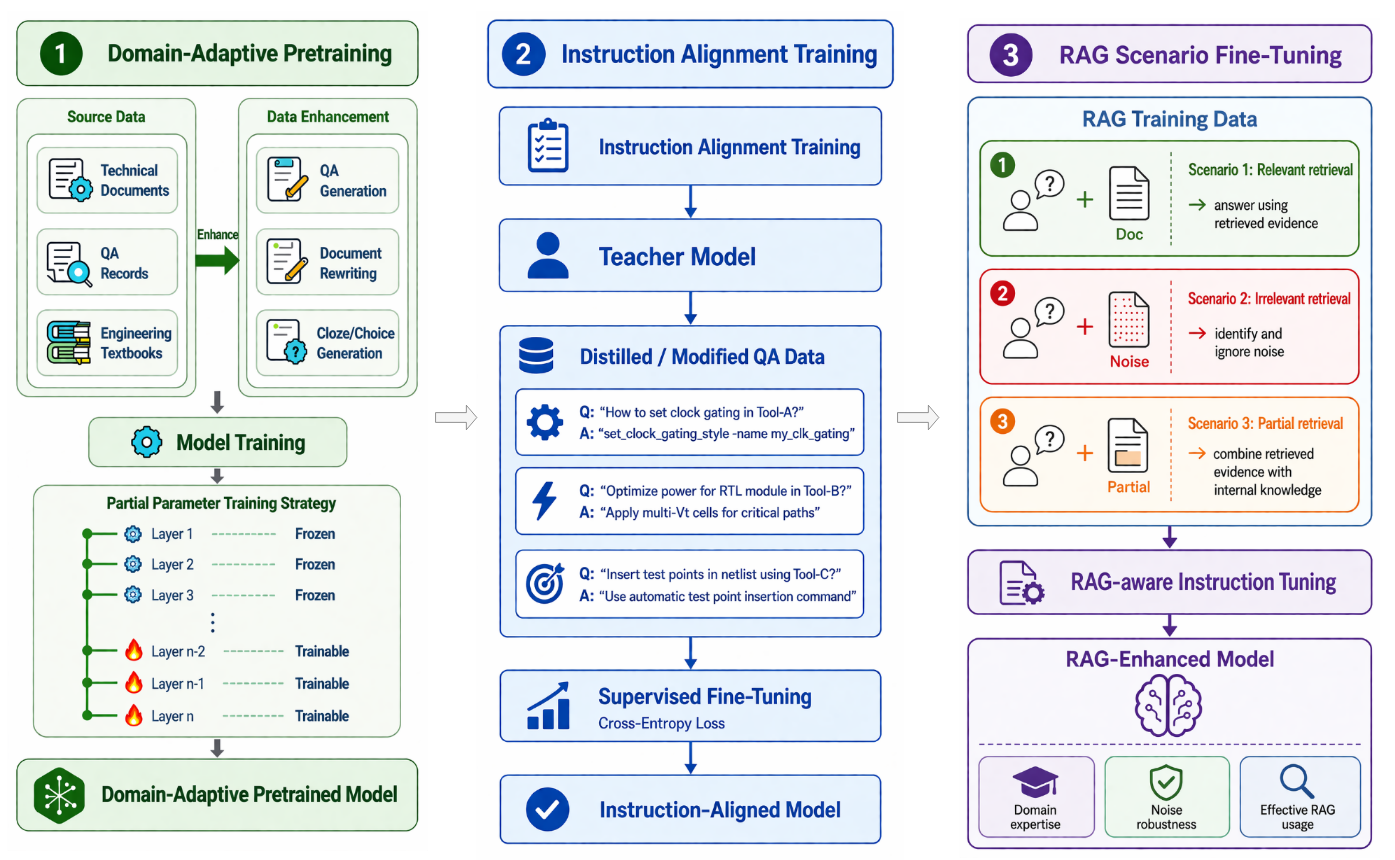}
    \caption{Overview of the ChipLingo three-stage training framework. Stage 1 (Domain-Adaptive Pretraining) constructs domain pretraining corpora through multi-source data fusion and augmentation strategies, employing partial parameter training to obtain a model with EDA domain expertise. Stage 2 (Instruction Alignment Training) leverages high-quality QA data distilled and rewritten from strong models, using supervised fine-tuning to enable instruction comprehension and task execution capabilities. Stage 3 (RAG Scenario Fine-Tuning) introduces training data covering various RAG scenarios including correct retrieval, irrelevant retrieval, and incomplete retrieval, enabling the model to achieve a balance between utilizing external knowledge and filtering noise.}
    \label{fig:overview}
\end{figure}

The overall architecture of the ChipLingo framework is illustrated in Figure~\ref{fig:overview}.

\subsection{Domain Data Preparation}

\begin{figure}[htbp]
    \centering
    \begin{subfigure}[b]{0.75\textwidth}
        \centering
        \includegraphics[width=\textwidth]{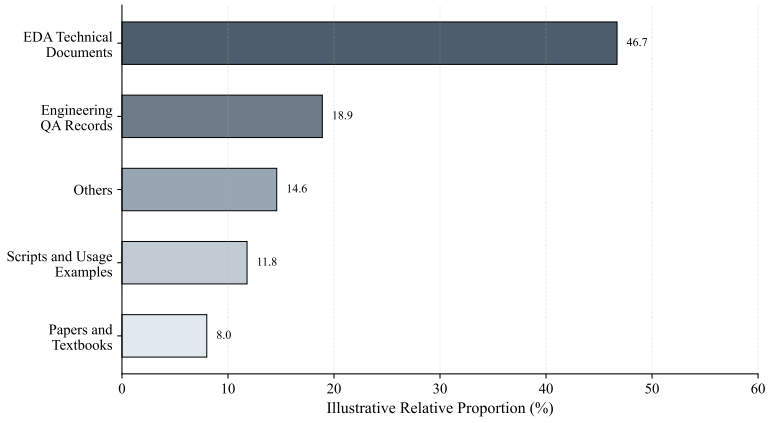}
        \caption{Source distribution of pretraining raw data}
        \label{fig:data-comp}
    \end{subfigure}
    \\[6pt]
    \begin{subfigure}[b]{0.75\textwidth}
        \centering
        \includegraphics[width=\textwidth]{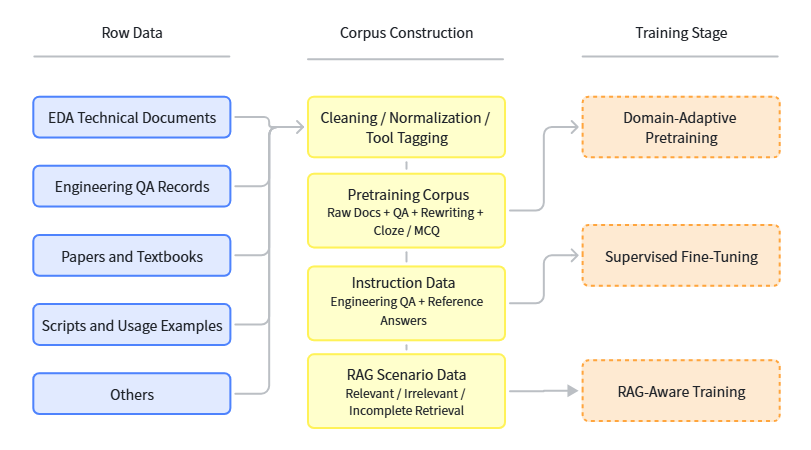}
        \caption{Organization from raw data to training corpora}
        \label{fig:data-pipe}
    \end{subfigure}
    \caption{ChipLingo data composition and training corpus organization}
    \label{fig:data}
\end{figure}

EDA domain data is characterized by dispersed sources, complex structures, and diverse formats. To construct high-quality training data, we systematically curated EDA documentation and engineering QA data, and expanded the training corpus through various data augmentation methods. The overall data composition and corpus organization are illustrated in Figure~\ref{fig:data}.

\subsubsection{Data Sources and Cleaning}

Training data primarily comes from the following categories:
\begin{itemize}
    \item EDA tool technical documentation
    \item Engineer question-answering records
    \item Technical papers and educational materials
    \item Tool usage examples and script documentation
\end{itemize}

The overall training corpus scale exceeds \textbf{200,000 pages of technical documentation}, covering \textbf{more than 50 EDA tools}, spanning multiple design stages including logic synthesis, physical implementation, simulation verification, and design-for-test analysis.

\subsubsection{Data Augmentation Strategies}

Relying solely on raw documentation data is often insufficient to form high-quality training corpora. Therefore, we explored multiple data augmentation methods:
\begin{enumerate}
    \item \textbf{QA Generation}: Automatically generating question-answer pairs from documentation.
    \item \textbf{Document Rewriting}: Semantically-preserving rewriting of document content to increase data diversity.
    \item \textbf{Cloze Generation}: Generating fill-in-the-blank tasks from technical descriptions.
    \item \textbf{Multiple Choice Generation}: Constructing training data in multiple-choice or single-choice question formats.
\end{enumerate}

Through comparative experiments on different augmentation strategies, we found that \textbf{QA-format data exhibits high efficiency for domain knowledge learning}, and thus constitutes a significant proportion of the training data.

\subsection{Domain-Adaptive Pretraining}

After completing data preparation, we perform domain-adaptive pretraining on the base model to enable it to acquire EDA-related knowledge.

\subsubsection{QA-Augmented Pretraining}

Traditional domain-adaptive pretraining typically uses only document text for training, with question-answering capabilities learned during subsequent supervised fine-tuning stages. However, in knowledge-intensive domains, this training approach may struggle to establish connections between knowledge and tasks.

Therefore, we introduce \textbf{QA-format data} during the pretraining phase, enabling the model to learn both domain knowledge and its application simultaneously. Experimental results indicate that, under the current experimental settings, incorporating QA data leads to improved performance on EDA tasks.

\subsubsection{Partial Parameter Training Strategy}

Domain-adaptive training often leads to degradation of model general capabilities. To achieve a balance between domain knowledge acquisition and general capability preservation, we explored a \textbf{partial parameter training strategy}. In this strategy, we freeze the parameters of several bottom layers and update only a selected subset of the remaining parameters, thereby reducing the impact of domain training on general capabilities.

\subsubsection{Limitations of Parameter-Efficient Fine-Tuning}

During experiments, we also compared the performance of full-parameter training and parameter-efficient fine-tuning methods on EDA tasks. Experimental results indicate that in knowledge-intensive tasks, parameter-efficient fine-tuning methods may underperform compared to full-parameter training.

A possible explanation is that EDA knowledge possesses highly complex structures and high information density, while low-rank updates may struggle to express such complex knowledge representations. Therefore, full-parameter training still maintains certain advantages in specific domain tasks.

\subsection{Instruction Alignment and RAG Training}

After completing domain-adaptive pretraining, we train the model's instruction comprehension capabilities through supervised fine-tuning and introduce retrieval scenario data to enhance the model's RAG capabilities.

\subsubsection{Instruction Data Construction}

The construction of instruction training data is a critical component in the ChipLingo framework that bridges domain knowledge and task execution capabilities. This process consists of three steps: raw QA collection, strong model distillation and rewriting, and quality filtering with formatting.

\textbf{Raw QA Collection.} We collected a large volume of engineer QA records from real engineering environments, covering multiple EDA tool scenarios including logic synthesis, physical implementation, simulation verification, and design-for-test. These raw records possess high domain value but often suffer from colloquial expressions, missing context, and incomplete answers, making them unsuitable for direct use in supervised fine-tuning.

\textbf{Strong Model Distillation and Rewriting.} To improve data quality, we utilized high-performance language models to distill and rewrite the raw QA data. Specifically, we input raw questions into strong models and requested structured reference answers that explicitly include reasoning processes. Additionally, we completed ambiguous question formulations and expanded overly brief answers, ensuring each sample contains clear question descriptions, complete reasoning chains, and accurate final answers. Through this step, the professional value of raw QA data is preserved while data format and expression quality are improved.

\textbf{Quality Filtering and Formatting.} We conducted multiple rounds of filtering on the distilled data, removing samples that were too short, unrelated to the EDA domain, or had low answer credibility. The final instruction dataset contains \textbf{approximately 40,000 high-quality samples}, with each sample uniformly formatted as a triple $(q_i, r_i, a_i)$, where $q_i$ is the question, $r_i$ is the reasoning process, and $a_i$ is the final answer. During instruction alignment training, we use only the question-answer pair $(q_i, a_i)$ as the supervised target, while $r_i$ is retained as an auxiliary annotation for quality control and analysis. This dataset is used not only for instruction alignment training but also provides foundational corpora for subsequent RAG scenario training.

\subsubsection{RAG Capability Degradation Phenomenon}

During experiments, we observed that after domain training, the model's RAG capabilities may exhibit notable degradation. Specifically, after acquiring domain knowledge, the model tends to rely more on parametric knowledge for answering while ignoring retrieved document information.

This phenomenon can be understood as the model's \textbf{parametric bias}. As domain training progresses, the model gradually acquires substantial domain knowledge, thus becoming more inclined to rely on internal knowledge when answering questions rather than utilizing external retrieval results. This leads to situations where model accuracy under correct retrieval context may actually be lower than under no-retrieval conditions.

\subsubsection{RAG Scenario Training Method}

To mitigate RAG capability degradation, we designed a set of \textbf{retrieval scenario training data}. This data simulates various RAG application scenarios:
\begin{itemize}
    \item \textbf{Correct Knowledge Retrieved}: The model learns to reference context for answer generation.
    \item \textbf{Irrelevant Knowledge Retrieved}: The model learns to ignore noise and rely on parametric knowledge.
    \item \textbf{Incomplete Retrieval Results}: The model learns to perform joint reasoning by combining context and parametric knowledge.
\end{itemize}

Algorithm~\ref{alg:rag-data} describes the construction process for RAG scenario training data.

\begin{algorithm}[htbp]
\caption{RAG Scenario Training Data Construction Algorithm}
\label{alg:rag-data}
\begin{algorithmic}[1]
\Require Domain QA dataset $D_{\text{qa}} = \{(q_i, a_i)\}_{i=1}^{N}$, retrieval corpus $\mathcal{R}$
\Ensure RAG scenario training dataset $D_{\text{rag}}$
\State $D_{\text{rag}} \gets \emptyset$
\For{each $(q, a) \in D_{\text{qa}}$}
    \State $C_{\text{rel}} \gets \text{Retrieve}(\mathcal{R}, q, k=3)$ \Comment{Retrieve relevant documents}
    \State $D_{\text{rag}} \gets D_{\text{rag}} \cup \{(q, C_{\text{rel}}, a)\}$ \Comment{Scenario 1: Correct knowledge}
    \State $C_{\text{irr}} \gets \text{SampleIrrelevant}(\mathcal{R}, q)$ \Comment{Sample irrelevant documents}
    \State $D_{\text{rag}} \gets D_{\text{rag}} \cup \{(q, C_{\text{irr}}, a)\}$ \Comment{Scenario 2: Irrelevant knowledge}
    \State $C_{\text{partial}} \gets \text{Subsample}(C_{\text{rel}}, \text{ratio}=0.5)$ \Comment{Subsample partial documents}
    \State $D_{\text{rag}} \gets D_{\text{rag}} \cup \{(q, C_{\text{partial}}, a)\}$ \Comment{Scenario 3: Incomplete knowledge}
\EndFor
\State \Return $D_{\text{rag}}$
\end{algorithmic}
\end{algorithm}

\subsection{Retrieval System Design}

During inference, ChipLingo integrates a retrieval system to obtain EDA-related documents. The retrieval system employs a hybrid retrieval strategy, combining semantic vector retrieval and keyword-based retrieval methods to improve recall. Additionally, the system selects the most relevant document segments through a document reranking mechanism and provides them as context input to the model for generating final answers.

Through this approach, the model can leverage both parametric knowledge and access external document information when needed.

\section{Experiments}
\label{sec:exp}

\subsection{Training Data and Implementation Details}

ChipLingo's training data primarily comes from EDA technical documentation, engineer QA records, and related technical materials. For data augmentation, we explored multiple approaches including document rewriting, QA generation, cloze generation, and multiple-choice generation. Experiments demonstrate that \textbf{combining multiple augmentation methods yields the best results}, with QA generation being particularly effective for improving domain question-answering capabilities.

In our experiments, we also observed that as training data scale increased from approximately 260k chunks to approximately 400k chunks, model performance on EDA tasks continued to improve, though with diminishing gains, indicating marginal returns on data scale for model performance.

\subsection{Evaluation Benchmark: EDA-Bench}

To evaluate model capabilities in EDA engineering scenarios, we constructed the \textbf{EDA-Bench} evaluation benchmark. EDA-Bench contains thousands of questions from real engineering scenarios, uniformly organized as short-answer questions. These questions primarily involve tool command usage, design flow comprehension, and common troubleshooting tasks.

The evaluation questions cover \textbf{four typical EDA tool categories} in chip design workflows:
\begin{itemize}
    \item \textbf{Logic Synthesis Tools}: Used for converting RTL descriptions to gate-level netlists;
    \item \textbf{Physical Implementation Tools}: Used for placement, routing, and timing optimization;
    \item \textbf{Simulation Verification Tools}: Used for functional verification and design behavior analysis;
    \item \textbf{Design-for-Test (DFT) Tools}: Used for testability design and test pattern generation.
\end{itemize}

\subsection{Evaluation Methodology}

EDA-Bench test questions are all short-answer format, making traditional string matching methods inadequate for accurate answer quality assessment. Therefore, we adopt the \textbf{LLM-as-a-Judge} automatic evaluation method, a paradigm that has been systematically studied and validated in works such as MT-Bench, Chatbot Arena, and G-Eval~\cite{zheng2023judge,liu2023geval}. Specifically, we use a high-performance language model as the evaluator to judge model-generated answers. Evaluator inputs include: the question, ground-truth answer, and model-predicted answer. The evaluation model's task is to determine whether the predicted answer is consistent with the ground-truth answer, outputting a \textbf{binary classification result of correct or incorrect}.

To verify the reliability of the automatic evaluation method, we conducted manual review on a subset of samples, and results indicate high consistency between automatic evaluation and human judgment.

\textbf{The related evaluation framework and dataset are currently undergoing curation and review, with plans for public release as an independent research contribution to promote the development of large language model research in the EDA domain.}

\subsection{Baseline Models}

In experiments, we compare ChipLingo with various general-purpose and industry-specific models. ChipLingo models are trained based on the \textbf{Qwen3 series models}~\cite{yang2025qwen3}, with two model scales: ChipLingo-8B and ChipLingo-32B. Baseline models include mainstream open-source large language models, strong open-source models, and industry-leading commercial models. All models are evaluated on the same EDA-Bench benchmark.

\subsection{Main Experimental Results}

\subsubsection{Overall Performance Comparison}

Table~\ref{tab:main-results} presents the overall performance of different models on EDA-Bench.

\begin{table}[htbp]
\centering
\caption{Overall performance of different models on EDA-Bench}
\label{tab:main-results}
\begin{tabular}{lc>{\columncolor{focusbg}}c}
\toprule
Model & Parameters & EDA-Bench Accuracy \\
\midrule
Qwen3-8B & 8B & 26.85\% \\
Qwen3-32B & 32B & 36.30\% \\
DeepSeek-v3.2 & 671B (37B active) & 56.28\% \\
ChipLingo-8B & 8B & 59.70\% \\
\rowcolor{bestbg}
ChipLingo-32B & 32B & \textbf{70.02\%} \\
GPT-5.4 & -- & 72.35\% \\
Claude-Sonnet-4.5 & -- & 71.11\% \\
\bottomrule
\end{tabular}
\end{table}

The proposed ChipLingo models demonstrate strong domain adaptability in semiconductor professional task evaluation. Specifically, ChipLingo-8B with only 8B parameters achieves 59.7\% accuracy, substantially outperforming the same-scale base model Qwen3-8B (26.85\%) and surpassing the larger general-purpose model DeepSeek-v3.2 (56.28\%). Meanwhile, ChipLingo-32B reaches 70.02\%, significantly improving over the same-scale general base model Qwen3-32B (36.30\%), with results approaching Claude-Sonnet-4.5 (71.11\%).

\subsubsection{Per-Tool Performance Analysis}

\begin{center}
    \includegraphics[width=0.85\textwidth]{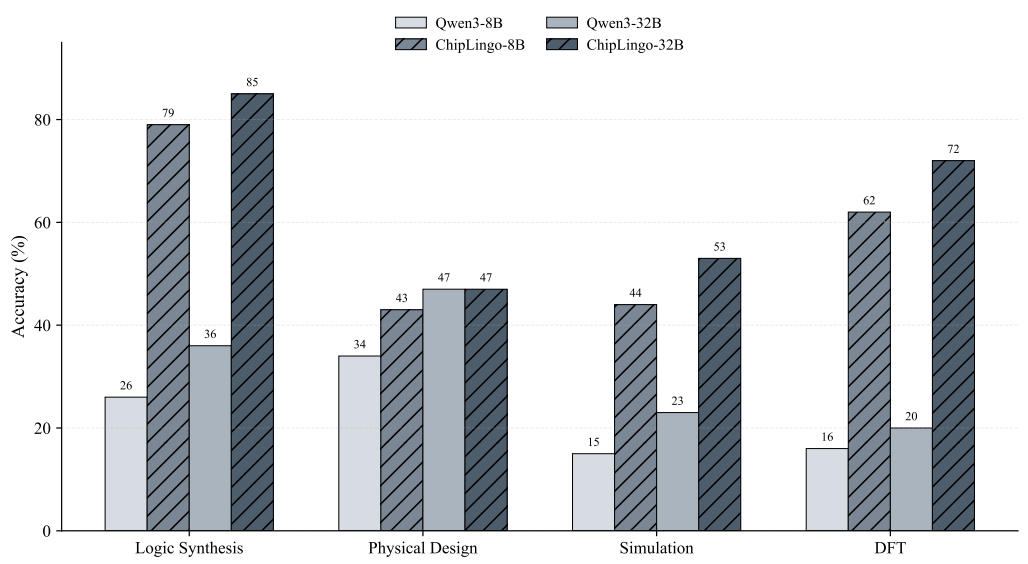}
    \captionof{figure}{Performance comparison across different EDA tool categories}
    \label{fig:tool-comp}
\end{center}

Figure~\ref{fig:tool-comp} presents representative evaluation set results extracted by tool category, illustrating relative performance trends across different tool scenarios. Results indicate that ChipLingo demonstrates overall superior performance across multiple tool task types, with the most notable improvements in DFT (Design-for-Test) related tasks. It should be noted that gains across different tool categories are not entirely uniform; in certain tool subsets, model improvements are relatively limited, suggesting that these scenarios still require more targeted corpus construction and training strategies.

\subsection{Ablation Studies}

To understand the effects of different training strategies, we conducted multiple ablation experiments. All ablation experiments below are based on Qwen3-8B.

\subsubsection{Effectiveness of QA-Augmented Pretraining}

\begin{center}
    \includegraphics[width=0.85\textwidth]{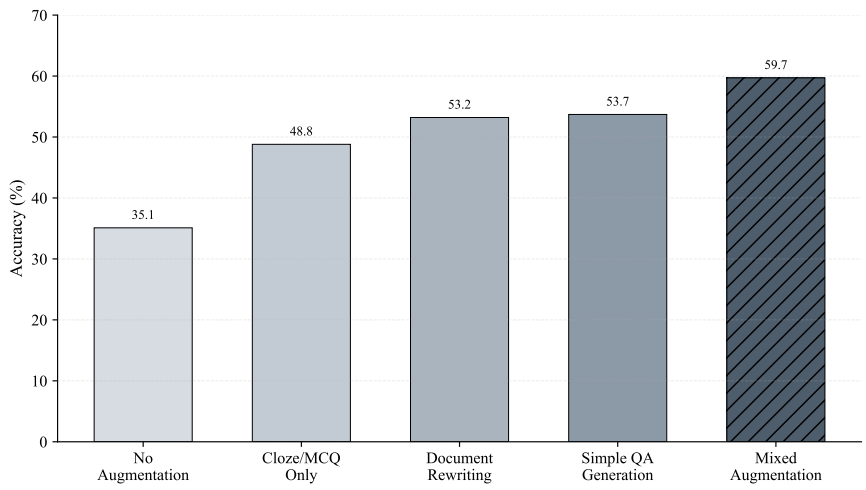}
    \captionof{figure}{Impact of different augmentation strategies on domain performance}
    \label{fig:aug}
\end{center}

Experimental results (Figure~\ref{fig:aug}) indicate that introducing multi-format constructed data during the pretraining phase helps improve the model's domain knowledge comprehension capabilities.

\begin{center}
    \includegraphics[width=0.85\textwidth]{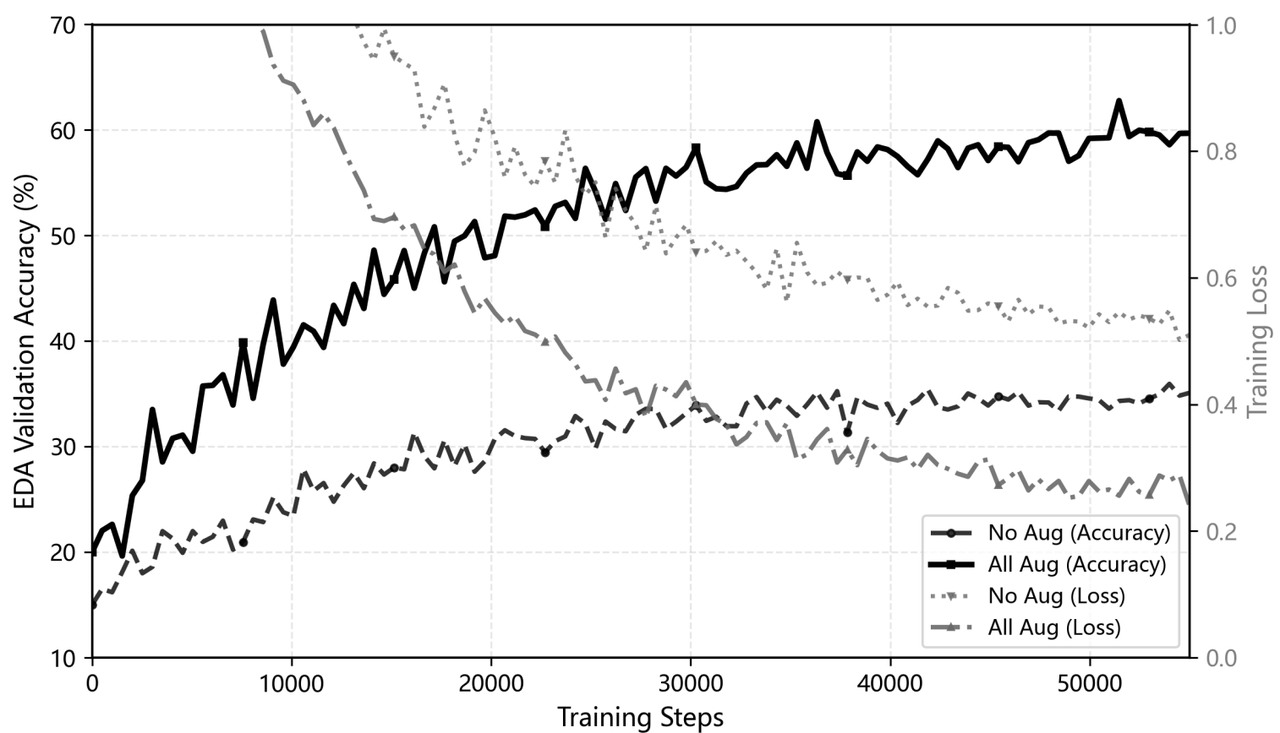}
    \captionof{figure}{Effect of multi-augmentation strategy pretraining on EDA domain learning}
    \label{fig:dynamics}
\end{center}

Comparing standard document pretraining and multi-augmentation strategy pretraining (All Aug) training dynamics over 55,000 training steps, the curves show that multi-augmentation strategies achieve higher final accuracy under current settings, while also exhibiting faster convergence and lower training loss (0.24 vs 0.51).

\subsubsection{Parameter Training Strategy Comparison}

\begin{table}[htbp]
\centering
\caption{Comparison of different parameter training strategies on EDA domain performance and general capability retention}
\label{fig:param}
\begin{tabular}{l>{\columncolor{focusbg}}c c c c c}
\toprule
\textbf{Method on Qwen3-8B} & \textbf{EDA-Bench} & \textbf{IFEval} & \textbf{SimpleQA} & \textbf{HumanEval} & \textbf{General Avg.} \\
\midrule
Base & 26.85 & 87.6 & 36.6 & 87.8 & 70.7 \\
LoRA & 46.8 & 85.2 & 32.4 & 83.6 & 67.1 \\
Full FT & \bestcell{61.0} & 82.1 & 28.7 & 79.2 & 63.3 \\
\rowcolor{tradebg}
Partial FT & 59.7 & 85.8 & 33.8 & 84.5 & 68.0 \\
\bottomrule
\end{tabular}
\end{table}

Table~\ref{fig:param} presents comparative results of different parameter training strategies on EDA-Bench and three general capability benchmarks. IFEval~\cite{zhou2023ifeval} evaluates instruction-following capability, SimpleQA~\cite{openai2024simpleqa} evaluates short-form factual question-answering capability, and HumanEval~\cite{chen2021humaneval} evaluates code generation capability. LoRA, while providing some domain performance improvement over the base model, still significantly underperforms full-parameter and partial-parameter training on EDA tasks. Full-parameter training achieves the highest accuracy on EDA-Bench, indicating stronger expressiveness for domain knowledge absorption; however, it simultaneously exhibits more significant performance degradation on IFEval, SimpleQA, and HumanEval. In comparison, partial-parameter training achieves results close to full-parameter training on EDA-Bench while maintaining better capability retention across all three general benchmarks. These results indicate that under current experimental settings, partial-parameter training demonstrates a more favorable empirical trade-off between domain adaptation effectiveness and general capability preservation.

\subsubsection{RAG Training Effect Analysis}

Table~\ref{tab:rag-degradation} presents quantified results of RAG capability degradation and recovery. For the base model Qwen3-8B, the relative improvement under the +correct retrieval condition is +7.3. After domain-adaptive pretraining (DAP), this value drops to $-5.5$, indicating that providing correct retrieval context actually degrades model performance. After subsequent supervised fine-tuning (+DAP+SFT), the relative improvement under the +correct retrieval condition remains negative at $-3.8$. After RAG scenario training, the relative improvement under the +correct retrieval condition recovers to +5.1, while performance loss under the +irrelevant retrieval condition improves from $-4.6$ to $-2.3$, indicating that model utilization of correct retrieval information is restored and demonstrates certain noise robustness under irrelevant retrieval conditions.

\begin{table}[htbp]
\centering
\caption{Quantified comparison of RAG capability degradation and recovery}
\label{tab:rag-degradation}
\begin{tabular}{lc>{\columncolor{focusbg}}c c}
\toprule
\textbf{Model} & \textbf{No Retrieval} & \textbf{+Correct Retrieval} & \textbf{+Irrelevant Retrieval} \\
\midrule
Qwen3-8B & 24.5\% & 31.8\%\,(+7.3) & 23.1\%\,($-1.4$) \\
+DAP only & 48.2\% & 42.7\%\,($-5.5$) & 43.0\%\,($-5.2$) \\
+DAP+SFT & 52.1\% & 48.3\%\,($-3.8$) & 47.5\%\,($-4.6$) \\
\rowcolor{bestbg}
+DAP+SFT+RAG & 59.7\% & \textbf{64.8\%\,(+5.1)} & \textbf{57.4\%\,($-2.3$)} \\
\bottomrule
\end{tabular}
\end{table}

\begin{center}
    \includegraphics[width=0.85\textwidth]{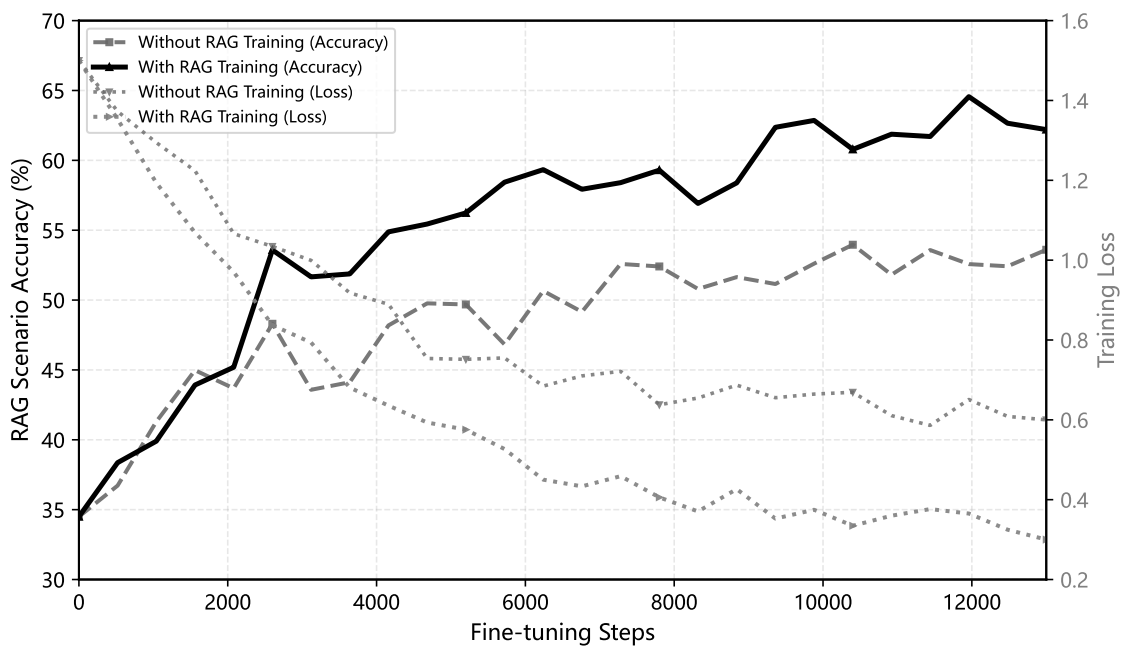}
    \captionof{figure}{RAG comparison during fine-tuning process}
    \label{fig:rag}
\end{center}

As shown in Figure~\ref{fig:rag}, without RAG training (+DAP+SFT), the model still exhibits negative gain under the +correct retrieval condition. After introducing RAG training with multiple scenario templates (+DAP+SFT+RAG), the model recovers positive gain under the +correct retrieval condition and shows better robustness under the +irrelevant retrieval condition. In addition, RAG training leads to a more stable convergence process.

\FloatBarrier
\subsection{Experimental Summary}

Based on comprehensive experimental results, we draw the following important conclusions:
\begin{enumerate}
    \item Under current experimental settings, domain-adaptive pretraining can improve model performance on EDA tasks.
    \item Experimental results indicate that introducing QA data during the pretraining phase helps enhance domain knowledge capabilities.
    \item Under current knowledge-intensive task settings, full-parameter training or partial-parameter training outperforms parameter-efficient fine-tuning methods such as LoRA.
    \item Multi-tool joint training can yield certain synergistic effects, but may also cause tool knowledge confusion.
    \item Experiments demonstrate that domain training may lead to RAG capability degradation, while explicit RAG scenario training helps restore this capability.
\end{enumerate}

These results support the effectiveness of the ChipLingo framework for EDA domain tasks.

\section{Analysis and Discussion}
\label{sec:analysis}

\subsection{Parameter Training Strategies and Knowledge Expressiveness}

During domain-adaptive pretraining, we systematically examined the impact of different parameter training strategies on model performance. Experimental results indicate that as the model absorbs EDA domain knowledge, its general capabilities often exhibit certain degrees of degradation, and parameter-efficient fine-tuning methods significantly underperform full-parameter training in such knowledge-intensive tasks. These observations reveal an inherent \textbf{capability competition} within the model's parameter space when simultaneously encoding general and domain-specific knowledge.

From a learning mechanism perspective, large language models must represent massive amounts of knowledge within a finite parameter space. When domain data occupies a large proportion, the model gradually allocates portions of its parameter resources to encode domain knowledge, thereby compressing the representation space for original general knowledge. Furthermore, EDA knowledge is characterized by abundant technical details, complex command structures, and fine-grained knowledge granularity, requiring the model to possess sufficient knowledge expressiveness. Methods such as LoRA rely on low-rank matrix updates to model weights, and their expressiveness may be insufficient to adequately capture such complex domain knowledge representations.

To mitigate these issues, we adopt a \textbf{partial parameter training strategy} that freezes the parameters of several bottom layers and updates only a selected subset of the remaining parameters, enabling the model to learn domain knowledge while preserving original language capabilities. Experimental results indicate that this strategy achieves performance close to full-parameter training. Based on comprehensive considerations of training efficiency, resource costs, and deployment feasibility, we ultimately adopt this approach for model training.

\subsection{Knowledge Transfer and Confusion in Multi-Tool Training}

A distinctive characteristic of the EDA domain is its complex tool ecosystem. Different tools typically correspond to different design stages while sharing certain foundational concepts, such as design constraints, timing analysis, and layout structures.

During experiments, we observed that \textbf{multi-tool training exhibits certain knowledge transfer effects}. When the model simultaneously learns multiple EDA tools of similar types, overall performance tends to improve. However, multi-tool training also introduces new challenges. In certain cases, the model uses commands from one tool to answer questions about another tool, manifesting as clear \textbf{tool knowledge confusion}.

\subsection{Causes and Remediation of RAG Degradation After Domain Training}

During experiments, we observed that after domain-adaptive training, the model's \textbf{RAG capability may exhibit degradation}. Specifically, even when the retrieval system returns documents containing correct answers, the model may still ignore this information and directly generate answers based on its parametric knowledge.

This phenomenon can be understood as the model's \textbf{parametric bias}. As domain training progresses, the model gradually acquires substantial domain knowledge, thus becoming more inclined to rely on internal knowledge when answering questions rather than utilizing external retrieval results. This manifests as negative relative improvement under +correct retrieval conditions, meaning that accuracy when provided with correct retrieval context is actually lower than accuracy under no-retrieval conditions.

To mitigate this issue, we further introduced diverse \textbf{RAG scenario training data} during the supervised fine-tuning stage, exposing the model to various input scenarios including correct retrieval, irrelevant retrieval, and incomplete retrieval during training. Experimental results indicate that this strategy can restore the model's ability to utilize external retrieval information to a certain extent: after RAG scenario training, the model's relative improvement under +correct retrieval conditions recovers from negative to positive values, while performance loss under +irrelevant retrieval conditions is reduced. This demonstrates that in knowledge-intensive domains, RAG capability does not naturally strengthen with domain knowledge training but requires targeted training mechanisms for maintenance. More importantly, such grounding capability is not only a requirement for traditional RAG question-answering modules but also serves as the foundation for subsequent EDA agent or harness systems to stably leverage documents, retrieval results, and tool feedback.

\section{Conclusion}
\label{sec:conclusion}

This paper presents \textbf{ChipLingo}, a large language model training pipeline combining domain-adaptive pretraining, instruction alignment training, and RAG scenario training, developed around knowledge-intensive Electronic Design Automation (EDA) question-answering scenarios. Experimental results demonstrate that this pipeline can improve model comprehension and question-answering performance for EDA-related knowledge.

To evaluate model performance on EDA tasks, we curated \textbf{EDA-Bench}, an evaluation benchmark. Experimental results indicate that under current evaluation settings, models with domain training show improved performance on EDA-Bench compared to general-purpose models. ChipLingo-8B outperforms some larger general-purpose models on this benchmark, while ChipLingo-32B achieves results approaching strong closed-source commercial models.

During training, we also observed several phenomena worthy of further investigation: under current experimental settings, QA augmentation contributes to improved domain performance, Partial FT demonstrates favorable empirical balance between domain adaptation effectiveness and selected general benchmark performance; furthermore, domain training may alter how models utilize external retrieval information, while explicit RAG scenario training helps mitigate this issue. These observations provide empirical references for understanding model adaptation in knowledge-intensive vertical domains, and also indicate that stable model capability for utilizing external knowledge is an important prerequisite for further constructing EDA agents and harness systems.

Building upon this work, we are continuing to expand ChipLingo's research and applications, gradually extending from single-turn question-answering and retrieval utilization to more complete harness capabilities, including more stable multi-step coordination combining external documents, tool feedback, and task workflows. Meanwhile, we will further expand EDA task coverage, continue scaling domain training data, and refine the \textbf{EDA-Bench} evaluation benchmark. \textbf{EDA-Bench is currently undergoing review and optimization, with plans for public release as an independent research contribution.}


\end{document}